# 3D-GANTex: 3D Face Reconstruction with StyleGAN3-based Multi-View Images and 3DDFA based Mesh Generation

Rohit Das
NTNU
Taipei City, Taiwan
61047086s@ntnu.edu.tw

Tzung-Han Lin
NTUST
Taipei City, Taiwan
thl@mail.ntust.edu.tw

Ko-Chih Wang
NTNU
Taipei City, Taiwan
kcwang@ntnu.edu.tw

**Abstract**

*Geometry and texture estimation from a single face image is an ill-posed problem since there is very little information to work with. The problem further escalates when the face is rotated at a different angle. This paper tries to tackle this problem by introducing a novel method for texture estimation from a single image by first using StyleGAN and 3D Morphable Models. The method begins by generating multi-view faces using the latent space of GAN. Then 3DDFA trained on 3DMM estimates a 3D face mesh as well as a high-resolution texture map that is consistent with the estimated face shape. The result shows that the generated mesh is of high quality with near to accurate texture representation.*

## 1. Introduction

The generation of photorealistic multi-view images from a single input remains a critical challenge in computer vision and graphics. While generative models have achieved significant progress in single-view face synthesis, extending this capability to multiple views is far more complex. This task involves reconstructing the 3D geometry and texture of a face from a 2D image, with additional challenges posed by variations in pose and the need to synthesize unseen areas. A particularly difficult subset of this problem is face frontalization, where the goal is to generate multiple views from a single frontal image. Key challenges in this process include 3D geometry recovery, inferring 3D information from 2D projections, texture extrapolation, maintaining viewpoint consistency, handling facial expressions and non-rigid deformations, managing occlusions, and achieving realism. While deep learning, image-based rendering, and 3D shape recovery methods have been proposed, the optimal solution remains elusive due to the inherent complexity of the task.

Texture generation and 3D model generation from a single image is another crucial aspect of multi-view face synthesis, presenting additional challenges such as limited texture information, perspective distortion, illumination variations, surface deformation, and texture resolution. Recent advances, particularly in generative adversarial networks (GAN)[1] have significantly improved texture generation. Techniques that rotate a 3D face mesh into arbitrary poses and render them back into 2D space [2, 3] have shown promise, but their effectiveness comes at the cost of significant computational resources due to the complexity of manipulating high-resolution 3D meshes. Research in this area continues to focus on overcoming the limitations of single-image texture synthesis through advanced neural networks and photorealistic rendering techniques. Despite progress, computational efficiency remains a major challenge, as the high fidelity of generated textures demands significant processing power, limiting the feasibility of real-time applications without substantial hardware support.

Given these advancements, several persistent challenges remain in 3D face generation, including synthesizing multi-view images from a single input, accurately estimating texture and 3D geometry from limited data, and obtaining suitable training datasets. To address these issues, the proposed paper introduces 3D-GANTex, a framework that generates front view image from any pose face image. The generated frontal image is then processed through a framework trained on morphable model 3DMM [4] which estimates the geometry as well as near to accurate texture. The method outperforms other models by not requiring labeled data, making it adaptable across various applications, from large-scale face recognition to avatar creation. Our contributions are as follows:

1. We propose 3D-GANTex, a novel inference pipeline that generates front face by embedding images in the latent space and estimates mesh and texture from parametric mesh models.
2. Our approach doesn't require prior information as we rely heavily on FFHQ [5] trained StyleGAN3 [6] and 3DDFAv2 [7] trained on 3DMM to address self-occlusion rather than detecting 2D landmarks.

The proposed method is an extension of Alaluf et al. [8] and 3DDFAv2 [7].

## 2. Related Work

**3D morphable face and head models.** The seminal work of Blanz and Vetter [4] was one of the first to introduce a model-based approach to represent variations in human faces using PCA on a set of 3D face scans. By adjusting the weights of these principal components, it is possible to generate new 3D faces that have similar characteristics to the faces in the original dataset, but with different shapes or textures.
The original 3DMM can be described as:

$$S = \overline{S} + A_{id}\alpha_{id} + A_{\exp}\alpha_{exp} \quad (1)$$

where $S$ is the 3D face mesh, $\overline{S}$ is the mean 3D shape, $\alpha_{id}$ is the shape parameter corresponding to the 3D shape base $A_{id}$, $A_{exp}$ the expression base and $\alpha_{exp}$ the expression parameter. After the 3D face is reconstructed, it can be projected onto the image plane with the scale orthographic projection:

$$V_{2d}(p) = f * Pr * R * (\overline{S} + A_{id}\alpha_{id} + A_{\exp}\alpha_{exp}) + t_{2d} \quad (2)$$

where $V_{2d}(p)$ is the projection function generating the 2D positions of model vertices, $f$ is the scale factor, $Pr$ is the orthographic projection matrix $\begin{pmatrix} 1 & 0 & 0 \\ 0 & 1 & 0 \end{pmatrix}$, $R$ is the rotation matrix constructed by Euler angles including pitch, yaw, roll and $t_{2d}$ is the translation vector. The complete parameters of 3DMM are $p = [f, R, t_{2d}, \alpha_{id}, \alpha_{\exp}]^T$.

Because of its linearity, 3DMM is very hard to align on non-rigid meshes. Also, the performance of these methods is restricted due to limitations of the 3D space defined by the face model basis or the templates [9, 10]. After the introduction of deep learning, various works focused on introducing non-linearity to face and head 3DMMs. Tran et al. [10] achieved a certain breakthrough by utilizing two CNN decoders instead of two PCA spaces, to learn nonlinearity from unlabeled images in a weakly-supervised manner. At the same time, graph convolutions [11, 12] and attention modules [12] have been proposed to model the head geometry.

**3DDFAV2:** 3DMM fitting methods are divided into 2 categories: the template fitting based [13, 14] and regression based [15, 16]. 3DDFAv2 [7] uses regression-based technique by introducing MobileNet[17] and to handle the optimization problem of the parameters regression framework, two different loss terms Weighted parameter Distance Cost (WPDC) and Vertex Distance Cost (VDC) were introduced. WPDC assigns different weights to each 3DMM parameter to account for their varying importance to accurately reconstruct the face VDC minimizes distance between ground truth vertices and predicted vertices. The meta-joint optimization looks ahead by k-steps with WPDC and VDC on the meta-train batches, then dynamically selects the better one according to the error on the meta-test batch. By doing so, the whole optimization converges faster and achieves better performance than the vanilla-joint optimization.

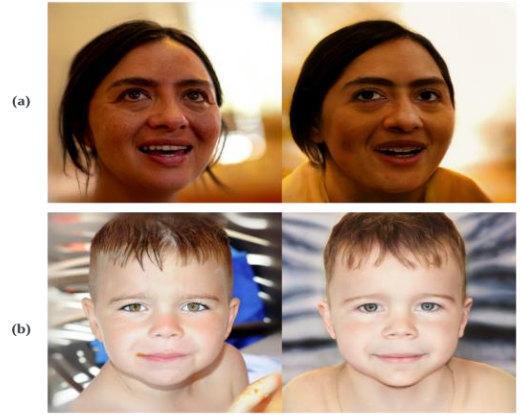

Figure 1: Generated images from ReStyle e4e encoder

**StyleGAN.** StyleGAN architectures [5, 6, 18, 19] are known for their semantically rich, disentangled and generally well-behaved latent spaces. These disentanglements really help in synthesizing exceedingly realistic images especially human face as it's been trained on FFHQ dataset [5] and enable editing [20-22] and image-to-image translation[23-25]. However, StyleGAN1[5] and StyleGAN2[18] affects the consistency and realism of generated images due to texture sticking phenomenon. For example, during interpolation within latent space, the hair and face typically do not move in unison. StyleGAN3[6] is designed to solve the problem and additionally offers translation and rotation equivariance.

To understand this phenomenon, it is important to understand StyleGAN3[6] functionalities and overall structure. First, in StyleGAN[5], a fully connected mapping network translates an initial latent code $z \sim N(0,1)^{512}$, into an intermediate code $w$ residing in a learned latent space $W$. Compared to StyleGAN2, StyleGAN3's synthesis network is composed of a fixed number of convolutional layers (16), irrespective of the output image resolution. They are denoted by $(w_0 \ldots w_{15})$ as the set of input codes passed to these layers. In StyleGAN3, the constant 4 x 4 input tensor from StyleGAN2 is replaced by Fourier features, that can be rotated and translated using four parameters $(sin_\alpha, cos_\alpha, x, y)$ aimed from from $w_0$ affine layer. The remaining layers are also fed $w_i$ into an independently learned affine layer, which yields modulation factors used to adjust the convolutional kernel weights. Since the translation and in-plane rotation of the synthesized images are given by explicit parameters obtained from $w_0$, the result may be easily adjusted by concatenating another transformation. We parameterize this transformation using three parameters $(r, t_x, t_y)$, where $r$ is the rotation angle (in degrees), and $t_x, t_y$ are the translation parameters, and denote the resulting image by:

$$y = G\left(w; (r, t_x, t_y)\right) \quad (3)$$

where, by default, $t_x = t_y = 0$ and $r = 0$. This transformation can be applied even in a generator trained solely on aligned data, enabling it to generate rotated and translated images.

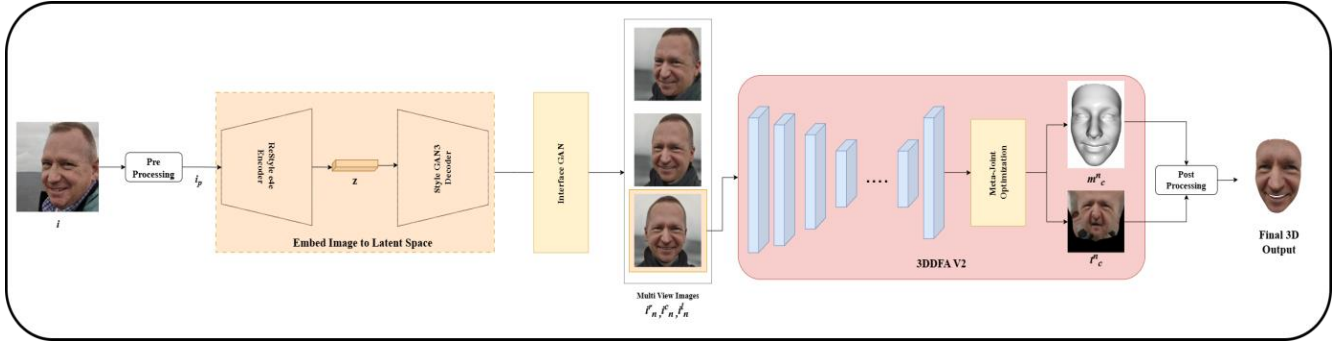

Figure 2: Method overview: Given an aligned image $i$, after preprocessing $i_p$ passes through e4e encoder where the residual scheme helps to retain high semantic details with very accurate latent code (left). With the latent code we pass it through InterfaceGAN to generate multi-view images with high fidelity ( $i_n^r$ , $i_n^c$ , $i_n^l$ ). After generating the front face, we pass it through 3DDFAv2 which after performing meta-joint optimization, estimates the mesh $m_c^n$ and texture $t_c^n$ thereby giving us the result output.

**Latent Space Embedding via Residual Encoders.** Encoders are crucial for retaining semantic information, particularly in face inversion tasks, which subsequently enable semantic editing. The goal is to retrieve a latent code $w$ that, when passed through a pre-trained StyleGAN generator, reproduces the original image. The most common method involves inverting the image into an extension of $W$, denoted $W_+$ [26]. To apply semantic edits on real image, we must invert the given image to StyleGAN's latent space. Previous works on learning-based inversion approaches and train encoders to map a given real image into its corresponding latent code [25, 27-29] have shown consistent results. Obtaining an accurate inversion is difficult, hence Alauf et al. [30] introduced a scheme which performs inversion using several forward passes by feeding the encoder with the output of the previous iteration (residual) along with the original input image. This allows the encoder to leverage knowledge learned in previous iterations to focus on relevant regions without losing much semantic information. We applied this ReStyle scheme to a modified e4e encoder [31], which can be adapted to various encoder architectures. The e4e encoder utilizes a Feature Pyramid Network [32] built on a ResNet [33] backbone. Unlike the original design, which extracts style features from three intermediate levels, our modified version extracts all style vectors from the final 16 × 16 feature map using a *map2style* block. This hierarchical encoder structure is particularly well-suited for well-structured domains such as faces. Figure 1 illustrates the e4e encoder in action, demonstrating its effectiveness in face inversion tasks.

**Face Rotation via Linear Latent Directions.** InterFaceGAN [34] is used for finding linear directions in $W$ for aligned and unaligned StyleGAN3 generators. Given a randomly sampled latent code $w \in W$, and editing direction $D$ and a step size $\delta$, the edited image is generated by $G_{aligned}(w + \delta D; (0,0,0))$ , where $G_{aligned}$ is the aligned generator. For unaligned images, there are two main approaches. The first is to use an unaligned generator, but this often leads to inaccurate attribute scores since classifiers are pre-trained on aligned images, resulting in poorly learned directions. The second approach is to use the aligned generator with user-defined transformations to control the image's rotation and position. This method allows the same latent directions to edit both aligned and unaligned images. The unaligned generator's latent space tends to be more entangled, especially for attributes like "smile." This entanglement is likely due to out-of-domain pseudo-aligned images and the challenge of achieving disentangled edits with linear directions. Given that the aligned generator consistently produces higher quality images (as demonstrated in StyleGAN3), we focus our analysis on using the aligned generator.

## 3. Methodology

Figure 2 shows the proposed pipeline for 3D-GANTex, in which $i$ refers to the input image and $i_p$ refers to the processed aligned image. To compute the pseudo transformation matrix $(r, t_x, t_y)$ for a given unaligned image $x_{unaligned}$, we went with dlib[35] for facial landmarks extraction. First, we detect the eyes from the dlib landmarks in both $x_{unaligned}$ and $x_{aligned}$ images. Then, we compute the rotation between these two sets by determining the angle between the lines connecting the eyes in both images. For translation, we rotate the aligned image and measure the vertical and horizontal distances between the left eye of the rotated aligned image and the unaligned image. These three values define the user-specified transformations that are passed to the Fourier features of StyleGAN3's synthesis network. Finally, the inversion and resulting reconstruction of the unaligned input $x_{unaligned}$ are given by:

$$w_{aligned} = E(x_{aligned}) \quad (4)$$

$$y_{unaligned} = G\left(w_{aligned}; (r, t_x, t_y)\right) \quad (5)$$

The e4e encoder allows precise control over generated images in the latent space $W_+$, enhancing image quality and flexibility. The Restyle scheme reinforces semantic details by applying residual style on the RGB channels of the generated image using the previously generated image. Subsequently, the latent codes of the encoder are combined with the residual latent code of the previous image, prioritizing originality and high-quality faces. Finally, the StyleGAN3 generator generates an image from this latent code.

We can only deduce this range by trial and error. After generating the disentangled latent code, it is processed by InterfaceGAN to edit the semantic details. Utilizing the computed pseudo transformation matrix $(r, t_x, t_y)$, we specify a minimum and maximum range for pose rotation. Explicit definition of this range is essential, as not all face poses share the same center position.

For instance, a 15° pose to the left may have a maximum range of -5, a minimum range of 3, with a central pose around 2. Conversely, a 45° pose might span from -3 to 4, with a central pose approximately at 2, as illustrated in Figure 4. Determining this range necessitates a trial-and-error approach.

To extract the 3D mesh $(m_c^n)$ and texture $(t_c^n)$ using 3DDFAv2, the identified front face from the multi-view images list is passed through the inference pipeline. This process extracts the 3DMM parameters, reconstructs the 3D face mesh, and maps the texture. Meta-joint optimization indirectly enhances the model's performance during training but is not utilized during inference. To predict the 3DMM parameters, the image is processed by MobileNet to predict $\alpha_{id}$ and $\alpha_{exp}$ and pose parameters. These parameters allow us to apply Equation *(1)* to obtain the required variables. Next, we employ the similarity transformation matrix to project the 3D mesh onto the image plane, as described in Equation *(2)*, ensuring alignment with the 2D image coordinates and matching the original image perspective. Following reconstruction, traditional texture mapping is performed, followed by post-processing refinements to achieve accurate texture.

| Method | SSIM↑ | MS-SSIM↑ | FSIM↑ | LPIPS↓ |
|---|---|---|---|---|
| FFHQ M-NG | **0.67** | **0.66** | 0.72 | 0.39 |
| FFHQ M | **0.67** | 0.62 | **0.78** | 0.36 |
| FFHQ F-NG | 0.60 | 0.62 | 0.75 | **0.24** |
| FFHQ F | 0.57 | 0.64 | 0.74 | 0.50 |
| ITW M-NG | **0.50** | 0.27 | 0.68 | 0.84 |
| ITW M | 0.49 | 0.23 | 0.65 | **0.48** |
| ITW F-NG | **0.50** | **0.36** | 0.69 | 0.71 |
| ITW F | 0.44 | 0.18 | **0.71** | 0.67 |

Table 1: Evaluation metrics for images generated from StyleGAN3 using the e4e encoder. The table includes results for the FFHQ (Flickr Faces High Quality) and ITW (In-the-Wild) datasets, categorized by gender (M- Male, F- Female) and the presence of glasses (NG)

## 4. Evaluation

We will discuss different metrics used to evaluate the 2D images generated from 3D-GANTex. Quantitative evaluations are mainly to evaluate the generated image quality and resemblance to the original image. For 3D and texture, we went ahead with qualitative evaluation. The proposed architecture was implemented and evaluated on a system equipped with an X64 based operating system and a 12th Gen Intel(R) Core (TM) i7-12700F (20 CPUs) processor running at ~2.1 GHz, coupled with 64.0 GB RAM and an NVIDIA GeForce RTX 4090 graphics card with 24 GB VRAM.

**Quantitative Evaluation.** For quantitative evaluation, we utilized the Structural Similarity Index (SSIM) [36], Multi-Scale Structural Similarity Index (MS-SSIM) [37], Feature Similarity Index (FSIM) [38], Learned Perceptual Image Patch Similarity (LPIPS) [39]. Table 1 represents the quantitative evaluation results for various image cases. We also assessed faces with glasses (G, NG) and in-the-wild images (ITW). The results indicate a high similarity score for FFHQ images, especially for faces without glasses, whereas in-the-wild images show lower similarity scores due to the challenge of accurately finding the latent space for unseen faces. This bias can be mitigated by training a neural network on a set of user images, as demonstrated in StyleGAN-ADA[19].

**Qualitative Evaluation.** For both qualitative and quantitative evaluation, we employed a 3D scanner[40]. Table 2 and Figure 3 represents the quantitative and qualitative evaluations, respectively. The Creality 3D-scanner demonstrates higher triangle consistency and texture due to its multi-view image acquisition, whereas our pipeline also achieves good triangle consistency and texture using only a single reference front image. While the CR-Scan 01 produces realistic textures, it is a time-consuming process, taking approximately 20 minutes to complete. In contrast, our pipeline significantly reduces this time, generating comparable textures in under 45 seconds. Future research could benefit from utilizing a different mesh dataset, such as FLAME[41] could be a better choice for more high-quality high triangle mesh datasets.

| 3D Mesh | No of Triangles | Average Triangle area from 50 vertices |
|---|---|---|
| 3D-GANTex | 76,073 | 3.47 |
| Creality CR-Scan 01 | 1,628,146 | 0.05 |

Table 2: Quantitative Analysis of 3D-GANTex v/s Creality CR-Scan 01

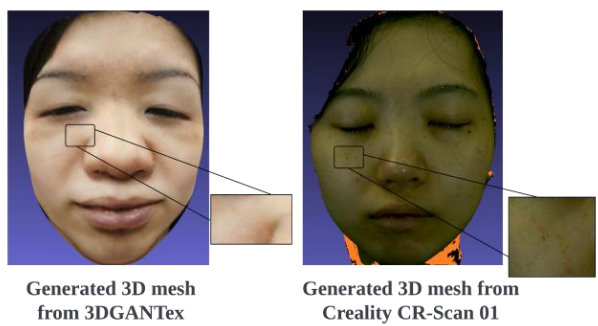

Figure 3: Texture difference between 3D-GANTex v/s CR-Scan 01

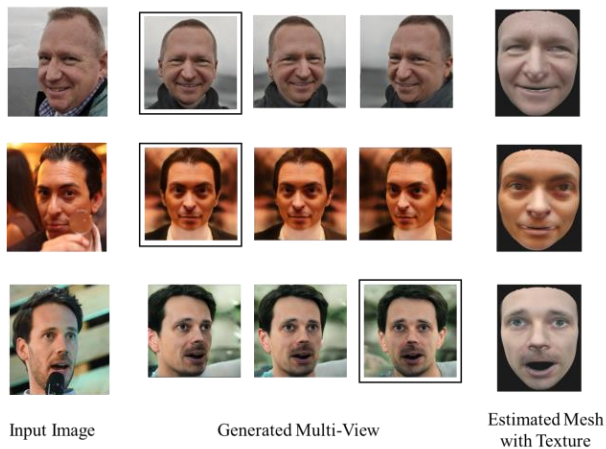

Figure 4: Few examples of 3DGANTex adapting to various pose information. The bordered images are termed as center pose

## 5. Conclusion

This work presents 3D-GANTex, a novel framework for generating high-quality 3D facial models from a single input image. We leverage the strengths of StyleGAN3 for multi-view image generation and 3DDFA for facial structure estimation and UV map generation. Our approach successfully embeds images into the latent space of StyleGAN3 using the Restyle-e4e encoder, enabling the generation of diverse views from a single input. We demonstrate the effectiveness of our method in capturing and preserving facial details through multi-view image generation and 3D model creation, particularly for images sourced from controlled datasets like FFHQ. However, our evaluation highlights limitations when dealing with real-world images, especially those featuring complex elements like eyeglasses. We aim to address these limitations and explore applications in real-world scenarios, including avatar generation.